\documentclass[11pt,a4paper]{article}
\usepackage[hyperref]{emnlp2020}
\usepackage{times}
\usepackage{latexsym}
\usepackage{pgfplots}
\usepackage{scalefnt}
\usepackage{amsfonts}
\usepackage{graphicx}
\usepackage{amsmath,bm}
\usepackage{mathrsfs}
\usepackage{multirow}
\usepackage{CJKutf8}
\usepackage{verbatim}
\usepackage{amssymb}
\usepackage{algorithm}
\usepackage{algorithmic}
\usepackage{tabu}
\usepackage{booktabs}
\usepackage{color}
\usepackage{subfigure}
\usepackage{url}

\usepackage{microtype}

\aclfinalcopy 

\title{SentiLARE: Sentiment-Aware Language Representation Learning with Linguistic Knowledge}

\author{Pei Ke\thanks{\quad Equal contribution}, Haozhe Ji$^*$, Siyang Liu, Xiaoyan Zhu, Minlie Huang\thanks{\quad Corresponding author}  \\
Department of Computer Science and Technology,
Institute for Artificial Intelligence, \\
State Key Lab of Intelligent Technology and Systems, \\ 
Beijing National Research Center for Information Science and Technology, \\
Tsinghua University, Beijing 100084, China \\
  {\tt kepei1106@outlook.com,\{jhz20,siyang-l18\}@mails.tsinghua.edu.cn} \\
  {\tt \{zxy-dcs,aihuang\}@tsinghua.edu.cn} \\}

\date{}

\begin{document}
\maketitle
\begin{abstract}

Most of the existing pre-trained language representation models neglect to consider the linguistic knowledge of texts, which can promote language understanding in NLP tasks. To benefit the downstream tasks in sentiment analysis, we propose a novel language representation model called SentiLARE, which introduces word-level linguistic knowledge including part-of-speech tag and sentiment polarity (inferred from SentiWordNet) into pre-trained models. We first propose a context-aware sentiment attention mechanism to acquire the sentiment polarity of each word with its part-of-speech tag by querying SentiWordNet. Then, we devise a new pre-training task called label-aware masked language model to construct knowledge-aware language representation. Experiments show that SentiLARE obtains new state-of-the-art performance on a variety of sentiment analysis tasks\footnote{The data, codes, and model parameters are available at \url{https://github.com/thu-coai/SentiLARE}.}.

\end{abstract}

\section{Introduction}

Recently, pre-trained language representation models such as GPT \cite{gpt12018,gpt22019}, ELMo \cite{elmo18}, and BERT \cite{bert19}
have achieved promising results in NLP tasks, including 
sentiment analysis \cite{bertpt19,hu2020dombert,yin2020sentibert}. These models capture contextual information from large-scale corpora via well-designed pre-training tasks. The literature has commonly reported that pre-trained models can be used as effective feature extractors and achieve state-of-the-art performance on various downstream tasks \cite{wang19glue}.

Despite the great success of pre-trained models, existing pre-training tasks like masked language model and next sentence prediction \cite{bert19} neglect to consider the linguistic knowledge. Such knowledge is important for some NLP tasks, particularly for sentiment analysis. For instance, existing work has shown that linguistic knowledge including part-of-speech tag \cite{qian15tagembedding,huang17syntactic} and word-level sentiment polarity \cite{linguisitic17} is closely related to the sentiment of longer texts. We argue that pre-trained models enriched with the linguistic knowledge of words will facilitate the understanding of the sentiment of the whole texts, thereby resulting in better performance on sentiment analysis.

There are two major challenges to construct knowledge-aware pre-trained language representation models which can promote the downstream tasks in sentiment analysis: 1) \textbf{Knowledge acquisition across different contexts}. Most of the existing work has adopted static sentiment lexicons as linguistic resource \cite{linguisitic17,chen2019sentence}, and equipped each word with a fixed sentiment polarity across different contexts. However, the same word may play different sentiment roles in different contexts due to the variety of part-of-speech tags and word senses.
2) \textbf{Knowledge integration into pre-trained models}.
Since the introduced word-level linguistic knowledge can only reflect the local sentiment role played by each word, it is important to deeply integrate knowledge into pre-trained models to construct sentence-level language representation, which can derive the global sentiment label of a whole sentence from local information.
How to build the connection between sentence-level language representation and word-level linguistic knowledge is underexplored.

In this paper, we propose a novel pre-trained language representation model called SentiLARE to deal with these challenges. First, to acquire the linguistic knowledge of each word, we label the word with its part-of-speech tag, and obtain the sentiment polarity via a context-aware sentiment attention mechanism over all the matched senses in SentiWordNet \cite{sentiwordnet10}.
Then, to incorporate linguistic knowledge into pre-trained models,
we devise a novel pre-training task called label-aware masked language model. This task involves two sub-tasks: 1) predicting the word, part-of-speech tag, and sentiment polarity at masked positions given the sentence-level sentiment label; 2) predicting the sentence-level label, the masked word and its linguistic knowledge including part-of-speech tag and sentiment polarity simultaneously.
We call the first sub-task \textit{early fusion} since the sentiment labels are aforehand integrated as input embeddings, whereas in the second sub-task, the labels are used as \textit{late supervision} to the model in the output layer.
These two sub-tasks are expected to establish the connection between sentence-level representation and word-level linguistic knowledge,
which can benefit downstream tasks in sentiment analysis.
Our contributions are in three folds:

\begin{itemize}
    \item We investigate the effectiveness of incorporating linguistic knowledge into pre-trained language representation models, and we reveal that injecting such knowledge via pre-training tasks can benefit various downstream tasks in sentiment analysis.
    
    \item We propose a novel pre-trained language representation model called SentiLARE. This model derives a context-aware sentiment polarity for each word using SentiWordNet, and adopts a pre-training task named label-aware masked language model to construct sentiment-aware language representations.
    
    
    \item We conduct extensive experiments on sentence-level and aspect-level sentiment analysis (including extraction and classification). Results show that SentiLARE obtains new state-of-the-art performance on a variety of sentiment analysis tasks.
    
\end{itemize}

\begin{figure*}[!htp]
  \centering
  \includegraphics[width=0.95\linewidth]{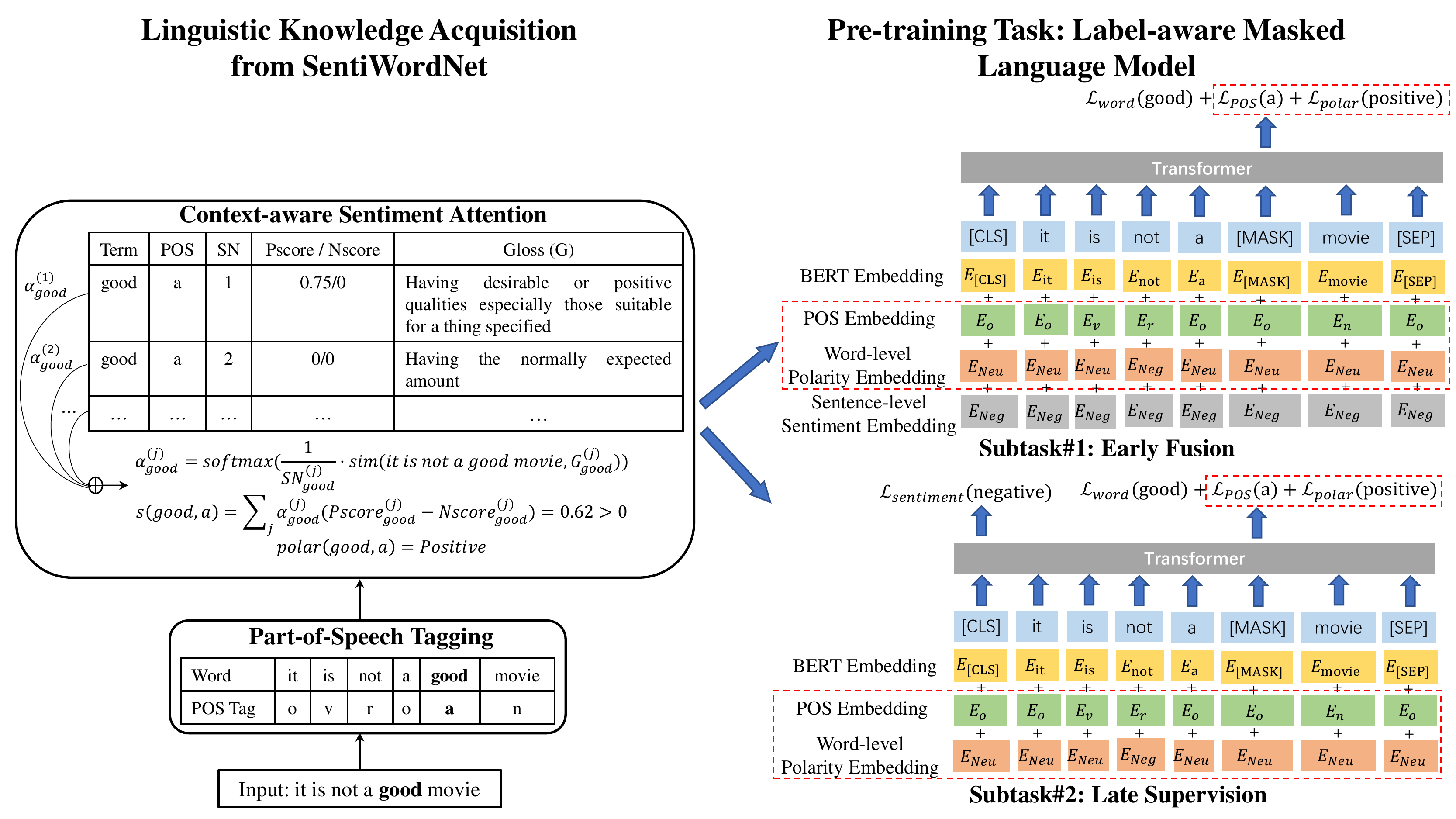}
  \caption{Overview of SentiLARE. This model first labels each word with its part-of-speech tag, and then uses the word and tag to match the corresponding senses in SentiWordNet. 
  The sentiment polarity of each word is obtained by weighting the matched senses with context-aware sentiment attention.
  During pre-training, the model is trained based on label-aware masked language model including \textit{early fusion} and \textit{late supervision}.
  Red dotted boxes denote that the linguistic knowledge is used in input embedding or pre-training loss function.
  }
  \label{fig:overview}
\end{figure*}

\section{Related Work}

\textbf{General Pre-trained Language Models}

\noindent Recently, pre-trained language representation models including ELMo \cite{elmo18}, GPT \cite{gpt12018,gpt22019}, and BERT \cite{bert19} become prevalent.
These models use LSTM \cite{lstm97} or Transformer \cite{transformer17} as the encoder to acquire contextual language representation, and explore
various pre-training tasks including masked language model and next sentence prediction \cite{bert19}.

Thanks to the great success of BERT on various NLP tasks, many variants of BERT have been proposed, which mainly fall into four aspects: 1) Knowledge enhancement: ERNIE-Tsinghua \cite{ernietsinghua19} / KnowBERT \cite{knowbert19} explicitly introduces knowledge graph / knowledge base to BERT, while ERNIE-Baidu \cite{erniebaidu19} designs entity-specific masking strategies during pre-training.
2) Transferability: TransBERT \cite{transferablebert19} conducts supervised post-training on the pre-trained BERT with transfer tasks to get a better initialization for target tasks.
3) Hyper-parameters: RoBERTa \cite{roberta19} measures the impact of key hyper-parameters to improve the under-trained BERT.
4) Pre-training tasks: SpanBERT \cite{spanbert19} masks consecutive spans randomly instead of individual tokens, while XLNet \cite{xlnet19} designs a training objective combining both reconstruction and autoregressive language modeling.


\noindent \textbf{Pre-trained Models for Sentiment Analysis}


\noindent Another line of work aims to build task-specific pre-trained models via post-training on the task data \cite{guru2020dontstop}.
For sentiment analysis, BERT-PT \cite{bertpt19} conducts post-training on the corpora which belong to the same domain of the downstream tasks
to benefit aspect-level sentiment analysis. 
DomBERT \cite{hu2020dombert} augments the training samples from relevant domains during the pre-training phase to enhance the performance on the aspect-level sentiment analysis of target domains. SentiBERT \cite{yin2020sentibert} devises a two-level attention mechanism on top of the BERT representation to capture phrase-level compositional semantics. 

Compared with the existing work on pre-trained models for sentiment analysis, our work integrates sentiment-related linguistic knowledge from SentiWordNet \cite{sentiwordnet10} into pre-trained models to construct knowledge-aware language representation, which can benefit a wide range of downstream tasks in sentiment analysis.

\noindent \textbf{Linguistic Knowledge for Sentiment Analysis}

\noindent Linguistic knowledge such as part of speech and word-level sentiment polarity is commonly used as external features in sentiment analysis. Part of speech is shown to facilitate the parsing of the syntactic structure of texts \cite{socher13pos}. It can also be incorporated into all layers of RNN as tag embeddings \cite{qian15tagembedding}. \citet{huang17syntactic} shows that part of speech can help to learn sentiment-favorable representations.

Word-level sentiment polarity is mostly derived from sentiment lexicons \cite{hu04lexicon,wilson05polarity}. \citet{priorpolarity13} obtains the prior sentiment polarity by weighting the sentiment scores over all the senses of a word in SentiWordNet \cite{sentiwordnet06,sentiwordnet10}.
\citet{Teng16lexicon} proposes a context-aware lexicon-based weighted sum model, which weights the prior sentiment scores of sentiment words to derive the sentiment label of the whole sentence. \citet{linguisitic17} models the linguistic role of sentiment, negation and intensity words via linguistic regularizers in the training objective of LSTM.
\section{Model}
\subsection{Task Definition and Model Overview}
Our task is defined as follows: given a text sequence $X=(x_1, x_2, \cdots, x_n)$ of length $n$ , our goal is to acquire the representation of the whole sequence $H=\left(h_1,h_2,\cdots,h_n\right)^{\top}\in\mathbb{R}^{n\times d}$ that captures the contextual information and the linguistic knowledge simultaneously, where $d$ indicates the dimension of representation vectors.

Figure \ref{fig:overview} shows the overview of our model, which consists of two steps: 1) Acquiring the part-of-speech tag and the sentiment polarity for each word; 2) Conducting pre-training via label-aware masked language model, which contains two pre-training sub-tasks, i.e., early fusion and late supervision.
Compared with existing BERT-style pre-trained models, our model enriches the input sequence with its linguistic knowledge including part-of-speech tag and sentiment polarity, and utilizes label-aware masked language model to capture the relationship between sentence-level language representation and word-level linguistic knowledge.

\subsection{Linguistic Knowledge Acquisition}
\label{sec:linguistic}

This module obtains the part-of-speech tag and the sentiment polarity for each word. The input of this module is a text sequence $X=(x_1,x_2,\cdots,x_n)$, where $x_i (1\leq i\leq n)$ indicates a word in the vocabulary.
First, our model acquires the part-of-speech tag $pos_i$ of each word $x_i$ via Stanford Log-Linear Part-of-Speech Tagger\footnote{\url{https://nlp.stanford.edu/software/tagger.html}}.
For simplicity, we only consider five POS tags including verb (\textit{v}), noun (\textit{n}), adjective (\textit{a}), adverb (\textit{r}), and others (\textit{o}).

Then, we acquire the word-level sentiment polarity $polar_i$ from SentiWordNet for each pair $(x_i,pos_i)$. In SentiWordNet, we can find $m$ different senses for the pair $(x_i, pos_i)$, each of which contains a sense number, a positive / negative score, and a gloss $(SN_{i}^{(j)},Pscore_{i}^{(j)}, Nscore_{i}^{(j)},G_{i}^{(j)})$, $1\leq j\leq m$, where $SN$ indicates the rank of different senses,
$Pscore / Nscore$ is the positive / negative score assigned by SentiWordNet, and $G$ denotes the definition of each sense.
Inspired by the existing work on inferring word-level prior polarity from SentiWordNet \cite{priorpolarity13} and unsupervised word sense disambiguation \cite{basile14leskwsd}, we propose a context-aware attention mechanism which simultaneously considers the sense rank and the context-gloss similarity to determine the attention weight of each sense:
\begin{align}
    \alpha_i^{(j)} = softmax(\frac{1}{SN_i^{(j)}}\cdot sim(X,G_i^{(j)})) \notag
\end{align}
where $\frac{1}{SN_i^{(j)}}$ approximates the impact of sense frequency because a smaller sense rank indicates more frequent use of this sense in natural language \cite{priorpolarity13}, and $sim(X,G_i^{(j)})$ denotes the textual similarity between the context and the gloss of each sense, which is commonly used as an important feature in unsupervised word sense disambiguation \cite{basile14leskwsd}. To calculate the similarity between $X$ and $G_i^{(j)}$, we encode them with Sentence-BERT (SBERT) \cite{reimers19sentencebert} which achieves the state-of-the-art performance on semantic textual similarity tasks, and obtain the cosine similarity between the vectors:
\begin{align}
    sim(X,G_i^{(j)}) = cos(\text{SBERT}(X),\text{SBERT}(G_i^{(j)})) \notag
\end{align}
%

Once we obtain the attention weight of each sense, we can calculate the sentiment score of each pair $(x_i,pos_i)$ by simply weighting the scores of all the senses:
%
\begin{align}
    s(x_i,pos_i) = \sum_{j=1}^{m} \alpha_i^{(j)} (Pscore_{i}^{(j)} - Nscore_{i}^{(j)}) \notag
\end{align}

Finally, 
the word-level sentiment polarity $polar_i$ for the pair $(x_i,pos_i)$ can be assigned with $Positive / Negative / Neutral$ when $s(x_i,pos_i)$ is positive / negative / zero, respectively. Note that if we cannot find any sense for $(x_i,pos_i)$ in SentiWordNet, $polar_i$ is assigned with $Neutral$.

\subsection{Pre-training Task}
\label{sec:pretraining}

Given the knowledge enhanced text sequence $X_{k}=\{(x_i,pos_i,polar_i)_{i=1}^n\}$, the goal of the pre-training task is to construct the knowledge-aware representation vectors $H=(h_1,\cdots,h_n)^{\top}$ which can promote the downstream tasks in sentiment analysis.
We devise a new supervised pre-training task called label-aware masked language model (LA-MLM), which introduces the sentence-level sentiment label $l$ into the pre-training phase to capture the dependency between sentence-level language representation and individual words.
It contains two separate sub-tasks: early fusion and late supervision.
%

\subsubsection{Early Fusion}

The purpose of early fusion is to recover the masked sequence conditioned on the sentence-level label, as shown in Figure \ref{fig:overview}. Assume that $\hat{X}_{k}$ denotes the knowledge enhanced text sequence with some masked tokens, we can obtain the representation vectors with the input of $\hat{X}_{k}$ and the sentence-level sentiment label $l$:
\begin{align}
    (h_{cls}^{EF},h_1^{EF},...,h_n^{EF},h_{sep}^{EF})=\text{Transformer}(\hat{X}_{k},l) \notag
\end{align}
where $h_{cls}^{EF}$ and $h_{sep}^{EF}$ are the hidden states of the special tokens \texttt{[CLS]} and \texttt{[SEP]}. The input embeddings of $\hat{X}_{k}$ contains the embedding used in BERT \cite{bert19}, the part-of-speech (POS) embedding and the word-level polarity embedding. Additionally, the embedding of the sentence-level sentiment label $l$ is \textit{early} added to the input embeddings. The model is required to predict the word, part-of-speech tag, and word-level polarity at the masked positions individually, thus the loss function is devised as follows:
\begin{align}
\mathcal{L}_{EF}& = -\sum_{t=1}^n m_t\cdot[\log P(x_t|\hat{X}_{k},l)+ \notag \\
&\log P(pos_t|\hat{X}_{k},l)+\log P(polar_t|\hat{X}_{k},l)] \notag
\end{align}
where $m_t$ is an indicator function and equals to $1$ iff $x_t$ is masked. The prediction probabilities $P(x_t|\hat{X}_{k},l)$, $P(pos_t|\hat{X}_{k},l)$ and $P(polar_t|\hat{X}_{k},l)$ are calculated based on the hidden state $h_t^{EF}$.
This sub-task explicitly exerts the impact of the global sentiment label on the words and the linguistic knowledge of words, enhancing the ability of our model to explore the complex connection among them.

\begin{table*} [!htp]
\centering
\small
\setlength{\tabcolsep}{1.0mm}{
\begin{tabular}{llc}
\hline
Task & Input Format & Output Hidden States \\
\hline
Sentence-level Sentiment Classification &   $\texttt{[CLS]} x_1, \cdots x_n \texttt{[SEP]}$ & $h_{\texttt{[CLS]}}$  \\
Aspect Term Extraction &   $\texttt{[CLS]} x_1 \cdots x_n \texttt{[SEP]}$ & $h_1,h_2,\cdots,h_n$ \\
Aspect Term Sentiment Classification & $\texttt{[CLS]} a_1 \cdots a_l \texttt{[SEP]} x_1 \cdots x_n \texttt{[SEP]}$ & $h_{\texttt{[CLS]}}$  \\
Aspect Category Detection &   $\texttt{[CLS]} x_1 \cdots x_n \texttt{[SEP]}$ & $h_{\texttt{[CLS]}}$ \\
Aspect Category Sentiment Classification & $\texttt{[CLS]} a_1 \cdots a_l \texttt{[SEP]} x_1 \cdots x_n \texttt{[SEP]}$ & $h_{\texttt{[CLS]}}$  \\
Text Matching &  $\texttt{[CLS]} x_1 \cdots x_n \texttt{[SEP]} y_1 \cdots y_m \texttt{[SEP]}$ & $h_{\texttt{[CLS]}}$  \\
\hline
\end{tabular}}
\caption{Fine-tuning setting of SentiLARE on downstream tasks. Both $x_1 \cdots x_n$ and $y_1\cdots y_m$ indicate the text sequences, while $a_1 \cdots a_l$ denotes the aspect term / category sequence. The output hidden states are then used in the classification / regression layer.}
\label{tab:finetuneinputoutput}
\end{table*}

\subsubsection{Late Supervision}
The late supervision sub-task aims to predict the sentence-level label and the word information based on the hidden states at \texttt{[CLS]} and masked positions respectively, as shown in Figure \ref{fig:overview}. Similar to early fusion, the representation vectors with the input of $\hat{X}_{k}$ are obtained as follows:
\begin{align}
    (h_{cls}^{LS},h_1^{LS},...,h_n^{LS},h_{sep}^{LS})=\text{Transformer}(\hat{X}_{k}) \notag
\end{align}
In this sub-task, the sentiment label $l$ is used as the \textit{late} supervision signal. Thus, the loss function to simultaneously predict the sentence-level sentiment label, words, and linguistic knowledge of words is shown as follows:
\begin{align}
\mathcal{L}_{LS} =& -\log P(l|\hat{X}_{k}) -\sum_{t=1}^n m_t\cdot[\log P(x_t|\hat{X}_{k})+ \notag \\
&\log P(pos_t|\hat{X}_{k})+\log P(polar_t|\hat{X}_{k})] \notag
\end{align}
%
where the sentence-level classification probability $P(l|\hat{X}_{k})$ is calculated based on the hidden state $h_{cls}^{LS}$. This sub-task enables our model to capture the implicit relationship among the sentence-level representation at \texttt{[CLS]} and word-level linguistic knowledge at masked positions.


Since the two sub-tasks are separate, we empirically set the percentage of pre-training data provided for the late supervision sub-task as 20\% and early fusion as 80\%.
As for the masking strategy, we increase the probability of masking words with positive / negative sentiment polarity from 15\% in the setting of BERT to 30\% because they are more likely to
impact the sentiment of the whole text.

\section{Experiment}

\subsection{Pre-training Dataset and Implementation}

We adopted the Yelp Dataset Challenge 2019\footnote{\url{https://www.yelp.com/dataset/challenge}} as our pre-training dataset. This dataset contains 6,685,900 reviews with 5-class review-level sentiment labels. Each review consists of 127.8 words on average.

Since our method can adapt to all the BERT-style pre-trained models, we used RoBERTa \cite{roberta19} as the base framework to construct Transformer blocks in this paper, and discussed the generalization ability to other pre-trained models like BERT \cite{bert19} in the following experiment.
The hyper-parameters of the Transformer blocks were set to be the same as RoBERTa-Base due to the limited computational power.
Considering the high cost of training from scratch, we utilized the parameters of pre-trained RoBERTa\footnote{\url{https://github.com/pytorch/fairseq}} to initialize our model. 
We also followed RoBERTa to use Byte-Pair Encoding vocabulary \cite{gpt22019} whose size was 50,265.
The maximum sequence length in the pre-training phase was 128, while the batch size was 400. We took Adam \cite{kingma2014adam} as the optimizer and set the learning rate to be 5e-5. The warmup ratio was 0.1. SentiLARE was pre-trained on Yelp Dataset Challenge 2019 for 1 epoch with label-aware masked language model, which took about 20 hours on 4 NVIDIA RTX 2080 Ti GPUs.

\subsection{Fine-tuning Setting}

We fine-tuned SentiLARE to the downstream tasks including sentence-level sentiment classification, aspect-level sentiment analysis, and general text matching tasks in our experiments.
We adopted the fine-tuning settings in the existing work \cite{bert19,bertpt19}, and showed the
input format and the output hidden states of each task in Table \ref{tab:finetuneinputoutput}. Note that the input embeddings in all the downstream tasks only contain BERT embedding, part-of-speech embedding and word-level polarity embedding. The hyper-parameters of fine-tuning on different datasets are reported in the Appendix.

\subsection{Baselines}

We compared SentiLARE with general pre-trained models, task-specific pre-trained models and task-specific models without pre-training.

\noindent \textbf{General Pre-trained Models}: We adopted BERT \cite{bert19}, XLNet \cite{xlnet19}, and RoBERTa \cite{roberta19} as general pre-trained baselines. These models achieve state-of-the-art performance on various NLP tasks.

\noindent \textbf{Task-specific Pre-trained Models}: We used BERT-PT \cite{bertpt19}, TransBERT \cite{transferablebert19}, and SentiBERT \cite{yin2020sentibert}
as task-specific pre-trained baselines.
Since TransBERT is not originally designed to deal with sentiment analysis tasks, 
we chose review-level sentiment classification on Yelp Dataset Challenge 2019 as the transfer task, and the downstream tasks in sentiment analysis as the target tasks.

\noindent \textbf{Task-specific Models without Pre-training}: We also chose some task-specific baselines without pre-training for corresponding tasks, including SC-SNN \cite{chen2019sentence}, DRNN \cite{wang18drnn}, ML \cite{sachan19ml} for sentence-level sentiment classification, DE-CNN \cite{xu18decnn} for aspect term extraction, CDT \cite{sun19cdt} for aspect term sentiment classification, TAN \cite{movahedi2019acd} for aspect category detection, and AS-Capsules \cite{wang2019ascapsules} for aspect category sentiment classification.

We evaluated all the pre-trained baselines based on the codes and the model parameters provided by the original papers.
For a fair comparison, all the pre-trained models were set to the base version, which possess a similar number of parameters (about 110M). The experimental results were presented with mean values over 5 runs. As for the task-specific baselines without pre-training, we re-printed the results on the corresponding benchmarks from the references.


\subsection{Sentence-level Sentiment Classification}

\begin{table} [!htp]
\centering
\small
\setlength{\tabcolsep}{1.5mm}{
\begin{tabular}{cccc}
\hline
\multirow{2}*{Dataset}  & Amount & \multirow{2}*{Length} & \multirow{2}*{\# classes} \\
 & (Train/Valid/Test) & & \\
\hline
SST & 8,544 / 1,101 / 2,210 & 19.2 &  5 \\
MR  & 8,534 / 1,078 / 1,050 & 21.7 &  2 \\
IMDB & 22,500 / 2,500 / 25,000 & 279.2  &  2  \\
Yelp-2 & 504,000 / 56,000 / 38,000 & 155.3 & 2 \\
Yelp-5 & 594,000 / 56,000 / 50,000 & 156.6 & 5 \\
\hline
\end{tabular}}
\caption{Statistics of sentence-level sentiment classification (SSC) datasets.}
\label{tab:sentenceleveldata}
\end{table}

We first evaluated our model on sentence-level sentiment classification benchmarks including
Stanford Sentiment Treebank (SST) \cite{socher13pos}, Movie Review (MR) \cite{pang05mr}, IMDB \cite{maas11imdb}, and Yelp-2/5 \cite{zhang15yelp}, which are widely used datasets at different scales. 
The detailed statistics of these datasets are shown in Table \ref{tab:sentenceleveldata}, which contain the amount of training / validation / test sets, the average length and the number of classes. Since MR, IMDB, and Yelp-2/5 don't have validation sets, we randomly sampled subsets from the training sets as the validation sets, and evaluated all the pre-trained models with the same data split.

\begin{table} [!htp]
\centering
\small
\setlength{\tabcolsep}{0.8mm}{
\begin{tabular}{l|c|c|c|c|c}
\hline
Model  & SST & MR & IMDB & Yelp-2 & Yelp-5 \\
\hline
SOTA-NPT & 55.20$^{\sharp}$ & 82.50$^{\sharp}$ & 93.57$^{\dagger}$ & 97.27$^{\ddagger}$ & 69.15$^{\ddagger}$ \\
\hline
BERT & 53.37 & 87.52 & 93.87 & 97.74 & 70.16 \\
XLNet & 56.33 & 89.45 & 95.27 & 97.41 & 70.23 \\
RoBERTa & 54.89 & 89.41 & 94.68 & 97.98 & 70.12 \\
BERT-PT & 53.24 & 87.30 & 93.99 & 97.77 & 69.90 \\
TransBERT & 55.56 & 88.69 & 94.79 & 96.73 & 69.53 \\
SentiBERT & 56.87 & 88.59 & 94.04 & 97.66 & 69.94 \\
\hline
\hline
SentiLARE & \textbf{58.59**} & \textbf{90.82**} & \textbf{95.71**} & \textbf{98.22**} & \textbf{71.57**} \\
\hline
\end{tabular}}
\caption{Accuracy on sentence-level sentiment classification (SSC) benchmarks (\%). SOTA-NPT means the state-of-the-art performance from the baselines without pre-training, where
the results marked with $\sharp$, $^{\dagger}$ and $^{\ddagger}$ are re-printed from \citet{chen2019sentence}, \citet{sachan19ml} and \citet{wang18drnn}, respectively. ** indicates that our model significantly outperforms the best pre-trained baselines on the corresponding dataset (t-test, \textit{p-value}$<0.01$).}
\label{tab:sentencelevel}
\end{table}

The results are shown in Table \ref{tab:sentencelevel}. We can observe that SentiLARE performs better than the baselines on the sentence-level sentiment classification datasets, thereby indicating the effectiveness of our knowledge-aware representation in sentiment understanding. Compared with vanilla RoBERTa, SentiLARE enhances the performance on all the datasets significantly. This demonstrates that for sentiment analysis tasks, linguistic knowledge can be used to enhance the state-of-the-art pre-trained model via the well-designed pre-training task.

\subsection{Aspect-level Sentiment Analysis}

\begin{table*} [!htp]
\centering
\scriptsize
\setlength{\tabcolsep}{1.0mm}{
\begin{tabular}{ccccccccc}
\hline
\multirow{2}*{Dataset} & \# sentences & \# terms & \# categories & \# sentiment & Amount of ATE & Amount of ATSC & Amount of ACD & Amount of ACSC \\
 & (Train/Test) & (Train/Test) & (Train/Test)  & classes & (Train/Valid/Test) & (Train/Valid/Test) & (Train/Valid/Test) & (Train/Valid/Test) \\
\hline
Lap14 & 3,045 / 800  & 2,358 / 654 & - & 3 & 1,338 / 150 / 422 & 2,163 / 150 / 638 & - & - \\
Res14 & 3,041 / 800 & 3,693 / 1,134 & 3,711 / 1,025 & 3 & 1,871 / 150 / 606 & 3,452 / 150 / 1,120 & 2,891 / 150 / 800 & 3,366 / 150 / 973 \\
Res16 & 2,000 / 676 & - & 2,507 / 859 & 3 & - & - & 1,850 / 150 / 676 & 2,150 / 150 / 751 \\
\hline
\end{tabular}}
\caption{Statistics of aspect-level sentiment analysis datasets, where ATE / ATSC / ACD / ACSC indicates aspect term extraction / aspect term sentiment classification / aspect category detection / aspect category sentiment classification, respectively.}
\label{tab:aspectleveldata}
\end{table*}

\begin{table*} [!h]
\centering
\small
\setlength{\tabcolsep}{1.0mm}{
\begin{tabular}{l|c|c|cc|cc|c|c|cc|cc}
\hline
Task & \multicolumn{2}{c|}{ATE} & \multicolumn{4}{c|}{ATSC} & \multicolumn{2}{c|}{ACD} & \multicolumn{4}{c}{ACSC} 
\\
\hline
Dataset  & Lap14 & Res14 & \multicolumn{2}{c|}{Lap14} & \multicolumn{2}{c|}{Res14} & Res14 & Res16 & \multicolumn{2}{c|}{Res14} & \multicolumn{2}{c}{Res16}  \\
\hline
Model & F1 &  F1  & Acc. & MF1.& Acc. & MF1.& F1 & F1 & Acc. & MF1. & Acc. & MF1. \\
\hline
SOTA-NPT & 81.59$^{\sharp}$ & - & 77.19$^{\dagger}$ & 72.99$^{\dagger}$ & 82.30$^{\dagger}$ & 74.02$^{\dagger}$ & 90.61$^{\ddagger}$ & 78.38$^{\ddagger}$ & 85.00$^{\flat}$ & 73.53$^{\flat}$ & - & - \\
\hline
BERT & 83.22 & 87.68 & 78.18 & 73.11 & 83.77 & 76.06 & 90.48 & 72.59 & 88.35 & 80.40 & 86.55 & 71.19 \\
XLNet & 86.02 & 89.41 & 80.00 & 75.88 & 84.93 & 76.70 & 91.35 & 73.00 & 91.63 & 84.79 & 87.46 & 73.06  \\
RoBERTa & 87.25 & 89.55 & 81.03 & 77.16 & 86.07 & 79.21 & 91.69 & 77.89 & 90.67 & 83.81 & 88.38 & 76.04 \\
BERT-PT & 85.99 & 89.40 & 78.46 & 73.82 & 85.86 & 77.99 & 91.89 & 75.42 & 91.57 & 85.08 & 90.20 & 77.09 \\
TransBERT & 83.62 & 87.88 & 80.06 & 75.43 & 86.38 & 78.95 & 91.50 & 76.27 & 91.43 & 85.03 & 90.41 & 78.56 \\
SentiBERT & 82.63 & 88.67 & 76.87 & 71.74 & 83.71 & 75.42 & 91.67 & 73.13 & 89.68 & 82.90 & 87.08 & 72.10 \\
\hline
\hline
SentiLARE & \textbf{88.22*} & \textbf{91.15**} & \textbf{82.16*} & \textbf{78.70*} & \textbf{88.32**} & \textbf{81.63**} & \textbf{92.22} & \textbf{80.71**} & \textbf{92.97**} & \textbf{87.30**} & \textbf{91.29} & \textbf{80.00} \\ 
\hline
\end{tabular}}
\caption{F1, accuracy (Acc.) and Macro-F1 (MF1.) on four aspect-level sentiment analysis tasks including aspect term extraction (ATE), aspect term sentiment classification (ATSC), aspect category detection (ACD) and aspect category sentiment classification (ACSC) (\%). 
SOTA-NPT means the state-of-the-art performance from the baselines without pre-training, where the results marked with $\sharp$, $\dagger$, $\ddagger$ and $\flat$ are re-printed from \citet{xu18decnn}, \citet{sun19cdt}, \citet{movahedi2019acd} and \citet{wang2019ascapsules}, respectively. - means that the results are not reported in the references. * indicates that our model significantly outperforms the best pre-trained baselines on the corresponding dataset (t-test, \textit{p-value}$<0.05$), while ** means \textit{p-value}$<0.01$.}
\label{tab:aspectlevel}
\end{table*}

Aspect-level sentiment analysis includes
aspect term extraction, aspect term sentiment classification, aspect category detection, and aspect category sentiment classification.
For aspect term based tasks, we chose SemEval2014 Task 4 for laptop (Lap14) and restaurant (Res14) domains \cite{pontiki14semeval} as the benchmarks, while for aspect category based tasks, we used SemEval2014 Task 4 for restaurant domain (Res14) and SemEval2016 Task 5 for restaurant domain (Res16) \cite{pontiki16semeval}. 
The statistics of these datasets are reported in Table \ref{tab:aspectleveldata}. We followed the existing work \cite{bertpt19} to leave 150 examples from the training sets for validation. Since the number of the examples with the \textit{conflict} sentiment label is rather small, we adopted the same setting as the existing work \cite{tang16aspect,bertpt19} and dropped these examples in the aspect term / category sentiment classification task.

We present the results of aspect-level sentiment analysis in Table \ref{tab:aspectlevel}. We can see that SentiLARE outperforms the baselines on all the four tasks, and most of the improvement margins are significant.
Interestingly, in addition to aspect-level sentiment classification, our model also performs well in aspect term extraction and aspect category detection. Since the aspect words are mostly nouns, part-of-speech tags may provide additional knowledge for the extraction task. In addition, the aspect terms can be signaled by neighboring sentiment words. This may explain why our knowledge-aware representation can help to extract (detect) the aspect term (category).

\subsection{Ablation Test}

To study the effect of the linguistic knowledge and the label-aware masked language model, we conducted ablation tests and presented the results in Table \ref{tab:ablation}. Since the two sub-tasks are separate, the setting of -EF / -LS indicates that the pre-training data were all fed into the late supervision / early fusion sub-task, and -EF-LS denotes that the pre-training task is changed from label-aware masked language model to vanilla masked language model, while the input embeddings still include part-of-speech and word-level polarity embeddings.
The -POS / -POL setting means that we removed the part of speech / word-level sentiment polarity in the input embeddings respectively, as well as in the supervision signals of the two sub-tasks. Obviously, -POS-POL indicates the complete removal of linguistic knowledge.

\label{sec:ablation}
\begin{table} [!h]
\centering
\small
\setlength{\tabcolsep}{0.55mm}{
\begin{tabular}{l|c|c|cc|c|cc}
\hline
Task & SSC & ATE & \multicolumn{2}{c|}{ATSC} & ACD & \multicolumn{2}{c}{ACSC} \\
\hline
Dataset & SST & Res14 & \multicolumn{2}{c|}{Res14} & Res16 & \multicolumn{2}{c}{Res14}  \\
\hline
Model & Acc. & F1 & Acc. & MF1. & F1 & Acc. & MF1.  \\
\hline
RoBERTa & 54.89 & 89.55 & 86.07 & 79.21 & 77.89 & 90.67 & 83.81  \\
\hline
\hline
SentiLARE & \textbf{58.59} & \textbf{91.15} & \textbf{88.32} & \textbf{81.63} & \textbf{80.71} & \textbf{92.97} & \textbf{87.30}  \\
\hline
 - EF & 58.44 & 90.82 & 87.70 & 81.11 & 80.42 & 92.70 & 86.42 \\
 - LS & 57.33 & 90.88 & 87.21 & 80.46 & 79.74 & 92.44 & 86.14 \\
 - EF - LS & 56.91 & 90.74 & 86.95 & 79.71 & 78.92 & 91.32 & 84.73 \\
\hline
 - POS & 58.15 & 90.94 & 87.98 & 81.38 & 80.27 & 92.51 & 86.61 \\
 - POL & 57.95 & 90.63 & 87.64 & 81.34 & 79.40 & 92.46 & 86.30 \\
 - POS - POL & 57.31 & 90.35 & 87.59 & 81.20 & 79.21 & 92.21 & 85.68 \\
\hline
\end{tabular}}
\caption{Ablation test on sentiment analysis tasks. 
EF / LS / POS / POL denotes early fusion / late supervision / part-of-speech tag / word-level polarity, respectively.}
\label{tab:ablation}
\end{table}

Results in Table \ref{tab:ablation} show that both the pre-training task and the linguistic knowledge contribute to the improvement over RoBERTa. Compared with early fusion, the late supervision sub-task plays a more important role in the classification tasks which depend on the global representation of the input sequence, such as SSC, ATSC, ACD and ACSC. Intuitively, the late supervision sub-task may learn a meaningful global representation at \texttt{[CLS]} by simultaneously predicting the sentence-level sentiment labels and the word knowledge. Thus, it contributes more to the performance on these classification tasks.

As for the impact of linguistic knowledge, the performance of SentiLARE degrades more in the setting of removing the word-level sentiment polarity. This implies that the word-level polarity can help the pre-trained model more to derive the global sentiment in the classification tasks and signal neighboring aspects in the extraction task.

\subsection{Analysis on Knowledge Acquisition}

To investigate whether our proposed context-aware knowledge acquisition method can help construct knowledge-aware language representation, we compared the context-aware sentiment attention described in \S\ref{sec:linguistic} with a context-free prior polarity acquisition algorithm \cite{priorpolarity13}. This algorithm simply acquires a fixed sentiment polarity for each word with its part-of-speech tag by weighting the sentiment score of each sense with the reciprocal of the sense number,
regardless of the variety of context.
All the other parts of SentiLARE remain unchanged for comparison between these two knowledge acquisition methods.

\begin{table} [!h]
\centering
\small
\setlength{\tabcolsep}{0.6mm}{
\begin{tabular}{l|c|c|cc|c|cc}
\hline
Task & SSC & ATE & \multicolumn{2}{c|}{ATSC} & ACD & \multicolumn{2}{c}{ACSC} \\
\hline
Dataset & SST & Res14 & \multicolumn{2}{c|}{Res14} & Res16 & \multicolumn{2}{c}{Res14}  \\
\hline
Model & Acc. & F1 & Acc. & MF1. & F1 & Acc. & MF1.  \\
\hline
SentiLARE & \multirow{2}*{\textbf{58.59}} & \multirow{2}*{\textbf{91.15}} & \multirow{2}*{\textbf{88.32}} & \multirow{2}*{\textbf{81.63}} & \multirow{2}*{\textbf{80.71}} & \multirow{2}*{\textbf{92.97}} & \multirow{2}*{\textbf{87.30}} \\
(w/ CASA) &  &  &  &  &  &  &  \\
\hline
SentiLARE & \multirow{2}*{56.86} & \multirow{2}*{90.54} & \multirow{2}*{88.05} & \multirow{2}*{81.26} & \multirow{2}*{79.00} & \multirow{2}*{92.72} & \multirow{2}*{86.59} \\
(w/ CFPP) &  &  &  &  &  &  &  \\
\hline 
\end{tabular}}
\caption{F1, accuracy (Acc.) and Macro-F1 (MF1.) of SentiLARE with context-aware sentiment attention (CASA) or context-free prior polarity (CFPP) on sentiment analysis tasks (\%).}
\label{tab:acquisition}
\end{table}

Results in Table \ref{tab:acquisition} show that our context-aware method performs better on all the tasks. This demonstrates that our context-aware attention mechanism can help SentiLARE model the sentiment roles of words across different contexts, thereby leading to better knowledge enhanced language representation.

\subsection{Analysis on Knowledge Integration}

To further demonstrate the importance of label-aware masked language model which deeply integrates linguistic knowledge into pre-trained models, we divided the test set of SST into two subsets according to the number of sentiment words (including positive and negative words determined by linguistic knowledge) in the sentences. Since there are 6.48 sentiment words on average in the sentences of SST's test set, we partitioned the test set into two subsets:  \textit{SST-Less} contains the sentences with no more than 7 sentiment words, and \textit{SST-More} includes the other sentences.
Intuitively and presumably, the sentences with more sentiment words may include more complex sentiment expressions.
We compared three models:  RoBERTa which does not use linguistic knowledge, SentiLARE-EF-LS which simply augments input embeddings with linguistic knowledge as described in \S\ref{sec:ablation}, and SentiLARE which deeply integrates linguistic knowledge via the pre-training task.

\begin{table} [!h]
\centering
\small
\setlength{\tabcolsep}{0.6mm}{
\begin{tabular}{l|c|c}
\hline
Model & SST-Less & SST-More  \\
\hline
RoBERTa & 55.61 & 53.52 \\
\hline
SentiLARE-EF-LS & 58.20 (+2.59) & 54.49 (+0.97) \\
SentiLARE & \textbf{59.72 (+4.11)} & \textbf{56.45 (+2.93)} \\
\hline 
\end{tabular}}
\caption{Accuracy of RoBERTa, SentiLARE-EF-LS, and SentiLARE on two subsets of SST's test set (\%).}
\label{tab:sentiwords}
\end{table}

Results in Table \ref{tab:sentiwords} show that SentiLARE-EF-LS already outperforms RoBERTa remarkably on SST-Less by simply augmenting input features with linguistic knowledge. However, on SST-More, SentiLARE-EF-LS only obtains marginal improvement over RoBERTa. As for our model, SentiLARE can consistently outperform RoBERTa and SentiLARE-EF-LS, and the margin between SentiLARE and SentiLARE-EF-LS is more evident on SST-More. This indicates that our pre-training task can help to integrate the local sentiment information reflected by word-level linguistic knowledge into global language representation, and facilitate the understanding of complex sentiment expressions.

\subsection{Analysis on Generalization Ability}

\textbf{Generalization to Other Pre-trained Models}: To study whether the introduced linguistic knowledge and the proposed pre-training task can improve the performance of other pre-trained models besides RoBERTa, we chose BERT \cite{bert19} and ALBERT \cite{lan20albert} as the base framework for evaluation. The experimental settings were set the same as SentiLARE based on RoBERTa.  

\begin{table} [!h]
\centering
\small
\setlength{\tabcolsep}{0.48mm}{
\begin{tabular}{l|c|c|cc|c|cc}
\hline
Task & SSC & ATE & \multicolumn{2}{c|}{ATSC} & ACD & \multicolumn{2}{c}{ACSC} \\
\hline
Dataset & SST & Res14 & \multicolumn{2}{c|}{Res14} & Res16 & \multicolumn{2}{c}{Res14}  \\
\hline
Model & Acc. & F1 & Acc. & MF1. & F1 & Acc. & MF1.  \\
\hline
BERT & 53.37 & 87.68 & 83.77 & 76.06 & 72.59 & 88.35 & 80.40    \\
\hline
SentiLARE & \multirow{2}*{\textbf{55.64}} & \multirow{2}*{\textbf{89.93}} & \multirow{2}*{\textbf{86.47}} & \multirow{2}*{\textbf{79.62}} & \multirow{2}*{\textbf{77.44}} & \multirow{2}*{\textbf{91.82}} & \multirow{2}*{\textbf{85.21}}    \\
(w/BERT) &  &  &  &  &  &  &  \\
\hline
\hline
ALBERT & 53.67 & 89.44 & 84.25 & 75.96 & 73.61 & 89.13 & 82.18 \\
\hline
SentiLARE & \multirow{2}*{\textbf{54.75}} & \multirow{2}*{\textbf{90.63}} & \multirow{2}*{\textbf{86.56}} & \multirow{2}*{\textbf{79.17}} & \multirow{2}*{\textbf{77.74}} & \multirow{2}*{\textbf{91.00}} & \multirow{2}*{\textbf{84.27}}    \\
(w/ALBERT) &  &  &  &  &  &  &  \\
\hline
\end{tabular}}
\caption{F1, accuracy (Acc.) and Macro-F1 (MF1.) of BERT, ALBERT, and SentiLARE based on BERT / ALBERT on sentiment analysis tasks (\%).
}
\label{tab:othermodels}
\end{table}

Results in Table \ref{tab:othermodels} show that SentiLARE based on BERT / ALBERT outperforms vanilla BERT / ALBERT on the datasets of all the tasks,
which demonstrates that our proposed method can adapt to different BERT-style pre-trained models to benefit the tasks in sentiment analysis.

\begin{table} [!htp]
\centering
\small
\setlength{\tabcolsep}{1.5mm}{
\begin{tabular}{cccc}
\hline
\multirow{2}*{Dataset}  & Amount & \multirow{2}*{Task Type} & \multirow{2}*{\# classes} \\
 & (Train/Valid/Test) & & \\
\hline
SCT & 1,771 / 100 / 1,871 & Classification & 2 \\
SICK  & 4,500 / 500 / 4,927 & Classification & 3 \\
STSb & 5,749 / 1,500 / 1,379 & Regression  & - \\
\hline
\end{tabular}}
\caption{Statistics of text matching datasets.}
\label{tab:textmatchingdata}
\end{table}

\noindent \textbf{Generalization to Other NLP Tasks}: Since sentiment is a common feature to improve the tasks of text matching \cite{cai17story,transferablebert19}, we chose three text matching tasks to explore whether our sentiment-aware representations can also benefit these tasks, including story ending prediction, textual entailment, and semantic textual similarity. We evaluated RoBERTa and SentiLARE on the datasets of SCT \cite{mostafazadeh16story}, SICK \cite{marelli14sick}, and STSb \cite{cer17stsb} for the three tasks respectively. The statistics of these datasets are reported in Table \ref{tab:textmatchingdata}. We followed the existing work \cite{transferablebert19} to preprocess the SCT dataset, and directly adopted the official version of SICK and STSb datasets.


\begin{table} [!htp]
\centering
\small
\setlength{\tabcolsep}{1.0mm}{
\begin{tabular}{l|c|c|cc}
\hline
Task & StoryEndPred & TextEntail & \multicolumn{2}{c}{SemTextSim} \\
\hline
Dataset & SCT & SICK & \multicolumn{2}{c}{STSb} \\
\hline
Model & Acc. & Acc. & Pear. & Spear. \\
\hline
RoBERTa & 90.95 & 88.22 & 88.37 & 87.61 \\
SentiLARE & \textbf{92.47} & \textbf{89.42} & \textbf{89.15} & \textbf{88.50} \\
\hline
\end{tabular}}
\caption{Accuracy (Acc.), Pearson (Pear.) and Spearman (Spear.) correlations of RoBERTa and SentiLARE on the tasks of story ending prediction, textual entailment, and semantic textual similarity (\%).}
\label{tab:story}
\end{table}

Results in Table \ref{tab:story} show that SentiLARE can enhance the performance of RoBERTa on these text matching tasks. This indicates that sentiment-related linguistic knowledge can be successfully integrated into the pre-trained language representation model to not only benefit the sentiment analysis tasks but also generalize to other sentiment-related NLP tasks. We will explore the generalization of our model to more NLP tasks as future work.


\section{Conclusion}

We present a novel pre-trained model called SentiLARE for sentiment analysis, which introduces linguistic knowledge from SentiWordNet via context-aware sentiment attention, and adopts label-aware masked language model to deeply integrate knowledge into BERT-style models through pre-training tasks. Experiments show that SentiLARE outperforms state-of-the-art language representation models on various sentiment analysis tasks, and thus facilitates sentiment understanding.


\section*{Acknowledgments}

This work was jointly supported by the NSFC projects (key project with No. 61936010 and regular project with No. 61876096), and the Guoqiang Institute of Tsinghua University with Grant No. 2019GQG1. This work was also supported by Beijing Academy of Artificial Intelligence, BAAI. We thank THUNUS NExT Joint-Lab for the support.

\bibliography{emnlp2020}
\bibliographystyle{acl_natbib}

\appendix

\section{Experimental Details}

\subsection{Hyper-parameter Setting}
\label{sec:hyperparamsearch}

We provided the hyper-parameter search space during pre-training in Table \ref{tab:pretrainsearch}. Grid search was used to select hyper-parameters, and the selection criterion was the classification accuracy on the validation set when we fine-tuned the pre-trained model on SST.

\begin{table} [!htp]
\centering
\small
\setlength{\tabcolsep}{1.0mm}{
\begin{tabular}{cc}
\hline
Hyper-parameter & Search Space \\
\hline
Percentage of Data in EF & \textit{choice}[0,0.2,0.4,0.6,0.8,1] \\
Masking Probability & \multirow{2}*{\textit{choice}[0.2,0.3,0.4,0.5]} \\
(Sentiment Words) & \\
Batch Size & \textit{choice}[400,512] \\
Training Epoch & \textit{choice}[1,2] \\
Learning Rate & 5e-5 \\
Warmup Ratio & 0.1 \\
Sequence Length & 128 \\
Maximum Gradient Norm & 1.0 \\
Optimizer & Adam \\
Epsilon (for Adam) & 1e-8 \\
\hline
\end{tabular}}
\caption{Hyper-parameter search space of SentiLARE during pre-training. \textit{choice} indicates that the listed numbers will be chosen with the same probability.}
\label{tab:pretrainsearch}
\end{table}

We also provided the detailed setting of hyper-parameters during fine-tuning on the datasets of sentiment analysis, including hyper-parameter search space in Table \ref{tab:finetunesearch} and best assignments in Table \ref{tab:finetunebest}. Note that we used HuggingFace's Transformers\footnote{\url{https://github.com/huggingface/transformers}} to implement our model, so all the hyper-parameters we reported were consistent with the codes of HuggingFace's Transformers. We utilized manual search to select the best hyper-parameters during fine-tuning. The number of hyper-parameter search trials for each dataset was 20. We used accuracy as our criterion for selection on all the sentiment analysis tasks except aspect term extraction and aspect category detection. For these two tasks, F1 was adopted as the selection criterion.

\begin{table} [!htp]
\centering
\small
\setlength{\tabcolsep}{0.6mm}{
\begin{tabular}{cc}
\hline
Hyper-parameter & Search Space \\
\hline
Learning Rate & \textit{choice}[1e-5,2e-5,3e-5,4e-5,5e-5] \\
Training Epoch & \textit{uniform-integer}[3,8] \\
Warmup Step & \textit{uniform-integer}[0,total\_step*0.2] \\
Batch Size & \textit{choice}[12,16,24,32] \\
Sequence Length & \textit{choice}[128,256,512] \\
Maximum Gradient Norm & 1.0 \\
Optimizer & Adam \\
Epsilon (for Adam) & 1e-8 \\
\hline
\end{tabular}}
\caption{Hyper-parameter search space of SentiLARE on the downstream sentiment analysis tasks. \textit{uniform-integer} means the integers in the interval can be selected uniformly. In the search space of warmup step, total\_step denotes the total training steps on different datasets.}
\label{tab:finetunesearch}
\end{table}

\begin{table*} [!htp]
\centering
\small
\setlength{\tabcolsep}{0.9mm}{
\begin{tabular}{c|c|c|c|c|c|c|c|c|c|c|c|c|c}
\hline
 Task & \multicolumn{5}{c|}{SSC} & \multicolumn{2}{c|}{ATE} & \multicolumn{2}{c|}{ATSC} & \multicolumn{2}{c|}{ACD} & \multicolumn{2}{c}{ACSC} \\
 \hline
 Dataset & SST & MR & IMDB & Yelp-2 & Yelp-5 & Lap14 & Res14 & Lap14 & Res14 & Res14 & Res16 & Res14 & Res16 \\
\hline
Learning Rate & 2e-5 & 3e-5 & 2e-5 & 2e-5 & 2e-5 & 3e-5 & 3e-5 & 3e-5 & 3e-5 & 3e-5 & 3e-5 & 3e-5 & 3e-5 \\
Training Epoch & 3 & 4 & 3 & 3 & 3 & 4 & 4 & 8 & 8 & 4 & 6 & 8 & 8 \\
Warmup Step & 100 & 20 & 100 & 12,600 & 8,500 & 0 & 0 & 0 & 150 & 100 & 0 & 0 & 60 \\
Batch Size & 12 & 24 & 24 & 12 & 12 & 12 & 12 & 16 & 16 & 16 & 16 & 16 & 32 \\
Sequence Length & 256 & 256 & 512 & 512 & 512 & 128 & 128 & 128 & 128 & 128  & 128 & 128 & 128 \\
\hline
\end{tabular}}
\caption{Best assignments of hyper-parameters on the sentiment analysis tasks.}
\label{tab:finetunebest}
\end{table*}

\subsection{Results on Validation Sets}

In addition to the performance on the test set of each dataset which has been reported in the main paper, we also provided the validation performance on the datasets of sentence-level and aspect-level sentiment analysis in Table \ref{tab:sentencelevelvalid}. As mentioned in Appendix \ref{sec:hyperparamsearch}, accuracy and F1 were used to select the best hyper-parameters, so we reported the validation performance of all the pre-trained models on these metrics.

\begin{table*} [!htp]
\centering
\small
\setlength{\tabcolsep}{1.0mm}{
\begin{tabular}{c|c|c|c|c|c|c|c|c|c|c|c|c|c}
\hline
Task & \multicolumn{5}{c|}{SSC} & \multicolumn{2}{c|}{ATE} & \multicolumn{2}{c|}{ATSC} & \multicolumn{2}{c|}{ACD} & \multicolumn{2}{c}{ACSC} \\
 \hline
Dataset & SST & MR & IMDB & Yelp-2 & Yelp-5 & Lap14 & Res14 & Lap14 & Res14 & Res14 & Res16 & Res14 & Res16 \\
 \hline
Model & Acc. & Acc. & Acc. & Acc. & Acc. & F1 & F1 & Acc. & Acc. & F1 & F1 & Acc. & Acc. \\
\hline
BERT & 52.77 & 87.05 & 93.72 & 97.77 & 70.56 & 81.62 & 83.48 & 80.67 & 77.33 & 86.96 & 72.12 & 84.67 & 86.00 \\
XLNet & 54.13 & 89.05 & 95.20 & 97.48 & 70.69 & 85.13 & 87.80 & 82.00 & 82.00 & 86.63 & 71.72 & 84.00 & 82.67 \\
RoBERTa & 54.13 & 89.71 & 94.68 & 98.06  & 70.65  & 88.75 & 88.09 & 84.67 & 82.67 & 89.21 & 77.62 & 86.00 & 85.33  \\
BERT-PT & 52.32 & 88.00  & 93.88 & 97.82 & 70.34  & 86.55 & 86.39 & 84.67 & 80.00 & 87.65 & 79.02 & 84.67 & 88.00  \\
TransBERT & 53.41 & 88.57 & 94.72 & 96.82 & 70.15 & 84.52 & 85.20 & 80.67 & 79.33 & 87.39 & 77.14 & 87.33 & 88.67 \\
SentiBERT & 54.50 & 88.76 & 94.12 & 97.83 & 70.44 & 83.23 & 83.48 & 82.67 & 81.33 & 86.90 & 73.65 & 84.67 & 84.00  \\
\hline
\hline
SentiLARE & 55.04 & 90.07 & 95.96 & 98.26 & 72.14 & 86.07 & 88.61 & 83.33 & 84.67 & 88.37 & 80.54 & 88.67 & 89.33 \\
\hline
\end{tabular}}
\caption{Accuracy (Acc.) and F1 on the validation sets of sentiment analysis benchmarks (\%).}
\label{tab:sentencelevelvalid}
\end{table*}

\subsection{Runtime}

The runtime of fine-tuning on different datasets of sentiment analysis was reported in Table \ref{tab:runtime}. We tested all the pre-trained models on 4 NVIDIA RTX 2080 Ti GPUs.

\begin{table*} [!htp]
\centering
\small
\setlength{\tabcolsep}{0.9mm}{
\begin{tabular}{c|c|c|c|c|c|c|c|c|c|c|c|c|c}
\hline
 Task & \multicolumn{5}{c|}{SSC} & \multicolumn{2}{c|}{ATE} & \multicolumn{2}{c|}{ATSC} & \multicolumn{2}{c|}{ACD} & \multicolumn{2}{c}{ACSC} \\
 \hline
 Dataset & SST & MR & IMDB & Yelp-2 & Yelp-5 & Lap14 & Res14 & Lap14 & Res14 & Res14 & Res16 & Res14 & Res16 \\
\hline
BERT & 22 & 16 & 42 & 1,389 & 1,492 & 5 & 5 & 15 & 23 & 9 & 9 & 10 & 9 \\
XLNet & 28 & 23 & 158 & 3,412 & 3,607 & 5 & 8 & 17 & 25 & 10 & 11 & 12 & 8 \\
RoBERTa & 23 & 18 & 65 & 1,674 & 1,930 & 5 & 6 & 16 & 26 & 9 & 8 & 10 & 9 \\
BERT-PT & 25 & 18 & 53 & 1,426 & 1,611 & 4 & 6 & 17 & 26 & 8 & 7 & 10 & 6 \\
TransBERT & 21 & 19 & 42 & 1,413 & 1,586 & 5 & 5 & 12 & 17 & 10 & 9 & 10 & 7 \\
SentiBERT & 16 & 16 & 53 & 1,336 & 1,503 & 3 & 4 & 9 & 13 & 7 & 5 & 8 & 5 \\
\hline
\hline
SentiLARE & 26 & 20 & 56 & 1,539 & 1,706 & 7 & 8 & 13 & 20 & 12 & 13 & 10 & 7 \\
\hline
\end{tabular}}
\caption{Average runtime of fine-tuning on different datasets for each pre-trained model (minutes).}
\label{tab:runtime}
\end{table*}

\section{Case Study}

\begin{figure}[!htp]
  \centering
  \includegraphics[width=1.0\linewidth]{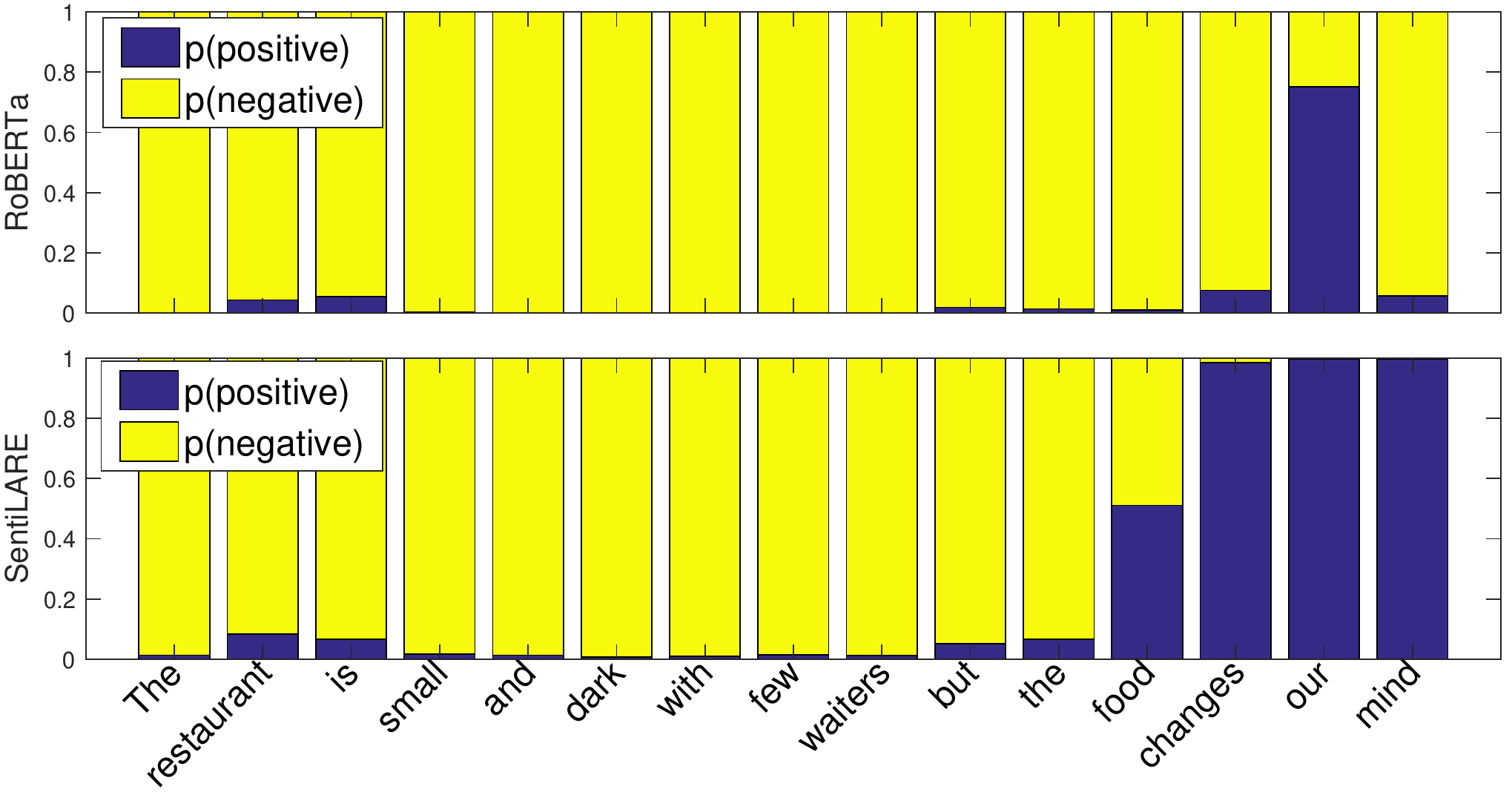}
  \caption{Visualization of the classification probability of RoBERTa and SentiLARE fine-tuned on Yelp-2. The bar indicates the output distribution of RoBERTa and SentiLARE with the input of the prefix subsequence at each position.}
  \label{fig:sentimentshift}
\end{figure}

To intuitively show that SentiLARE can integrate linguistic knowledge into pre-trained models to promote sentiment analysis, we provided a case, and visualized the classification probability of all the prefix subsequences truncated at each position in Figure \ref{fig:sentimentshift}. For example, the prediction probability at the position of \textit{small} was acquired by the hidden state of [CLS] obtained by RoBERTa / SentiLARE with the input ``[CLS] The restaurant is small [SEP]''. Both RoBERTa and SentiLARE were fine-tuned on the Yelp-2 dataset.

Compared with RoBERTa, our model enhanced with word-level linguistic knowledge can successfully capture the sentiment shift caused by the word \textit{change} in this sentence, thereby determining the correct sentence-level sentiment label.

\section{Analysis on Textual Similarity in Knowledge Acquisition}

\begin{table} [!h]
\centering
\small
\setlength{\tabcolsep}{0.48mm}{
\begin{tabular}{l|c|c|cc|c|cc}
\hline
Task & SSC & ATE & \multicolumn{2}{c|}{ATSC} & ACD & \multicolumn{2}{c}{ACSC} \\
\hline
Dataset & SST & Res14 & \multicolumn{2}{c|}{Res14} & Res16 & \multicolumn{2}{c}{Res14}  \\
\hline
Model & Acc. & F1 & Acc. & MF1. & F1 & Acc. & MF1.  \\
\hline
SentiLARE & \multirow{2}*{\textbf{58.59}} & \multirow{2}*{\textbf{91.15}} & \multirow{2}*{\textbf{88.32}} & \multirow{2}*{\textbf{81.63}} & \multirow{2}*{\textbf{80.71}} & \multirow{2}*{\textbf{92.97}} & \multirow{2}*{\textbf{87.30}} \\
(w/SBERT) &  &  &  &  &  &  &  \\
\hline
SentiLARE & \multirow{2}*{57.41} & \multirow{2}*{90.66} & \multirow{2}*{88.10} & \multirow{2}*{81.27} & \multirow{2}*{80.07} & \multirow{2}*{92.75} & \multirow{2}*{86.87} \\
(w/WordVec) &  &  &  &  &  &  &  \\
\hline 
\end{tabular}}
\caption{F1, accuracy (Acc.) and Macro-F1 (MF1.) of SentiLARE with Sentence-BERT (SBERT) or word vectors (WordVec) on sentiment analysis tasks (\%).}
\label{tab:similarityacquisition}
\end{table}

Since Sentence-BERT is costly to calculate the textual similarity between contexts and glosses, we compared it with another lighter textual similarity algorithm \cite{basile14leskwsd} which computes the representation vectors of sentences by averaging the embedding vectors of their constituent words. We used 300-dimensional GloVe\footnote{\url{https://nlp.stanford.edu/projects/glove/}} as the word vectors to obtain textual similarity.

Results in Table \ref{tab:similarityacquisition} show that Sentence-BERT performs better on all the tasks. Nevertheless, static word vectors are more computationally efficient in linguistic knowledge acquisition with acceptable performance drop.

\end{document}